\pdfoutput=1

\documentclass[11pt]{article}

\usepackage{ACL2023}

\usepackage{times}
\usepackage{latexsym}
\usepackage{hyperref}

\usepackage[T1]{fontenc}

\usepackage[utf8]{inputenc}

\usepackage{microtype}

\usepackage{inconsolata}

\usepackage{graphicx} %
\usepackage{booktabs}

\usepackage{amsmath,amssymb,amsfonts}
\usepackage{bm}
\usepackage{amsthm}
\usepackage{algorithm}
\usepackage{algorithmic}
\title{Improving Empathetic Dialogue Generation by Dynamically Infusing Commonsense Knowledge}

\author{Hua Cai$^{* \dagger}$ \ Xuli Shen$^{*}$ \ Qing Xu \ Weilin Shen \ Xiaomei Wang  \\ {\bf Weifeng Ge \ Xiaoqing Zheng \ Xiangyang Xue} \\ UniDT Technology, Shanghai, China \\
        School of Computer Science, Fudan University, Shanghai, China }

\begin{document}
\maketitle
\begin{abstract}
In empathetic conversations, individuals express their empathy towards others. Previous work has mainly focused on generating empathetic responses by utilizing the speaker’s emotion. Besides, external commonsense knowledge has been applied to enhance the system's understandings of the speaker's situation. However, given an event, commonsense knowledge base contains various relations, potentially leading to confusion for the dialogue system. Consequently, inconsistencies arise among the emotion, generated response and speaker's contextual information. To this end, we propose a novel approach for empathetic response generation, which incorporates an adaptive module for commonsense knowledge selection to ensure consistency between the generated empathetic responses and the speaker's situation. This selected knowledge is used to refine the commonsense cognition and empathy expression for generated responses. Experimental results show that our approach significantly outperforms baseline models in both automatic and human evaluations, exhibiting the generation of more coherent and empathetic responses. Moreover, case studies highlight the interpretability of knowledge selection in the responses and the effectiveness of adaptive module in our model. Code:  \href{https://github.com/Hanscal/DCKS}{https://github.com/Hanscal/DCKS}.
{\let\thefootnote\relax\footnote{{$^{*}$These authors contributed equally to this work.}}}
{\let\thefootnote\relax\footnote{{{$^{\dagger}$Corresponding author: Hua Cai (hua.cai@unidt.com) }}}}
\end{abstract}

\section{Introduction}

Empathy is a desirable human ability in our daily conversations. It is known as a complex multi-dimensional construct encompassing social, cognitive, and emotional processes, which enables us to experience the emotion of others through various emotional stimuli and to understand the implicit mental states of others \cite{Davis1983,Zheng:CoMAE}.  Previous research \cite{Rashkin:2019,Lin:2019,Majumder:2020,Li:2021} has been conducted on dialogue systems to enhance its empathy ability in open-domain. In order to generate empathetic responses, one line of growing interests is incorporating commonsense knowledge into conversation modeling \cite{ghosal2020cosmic,Zhou:2021,Sabour:2021}. 

\begin{figure}[t]
\centering
\includegraphics[width=0.95\columnwidth]{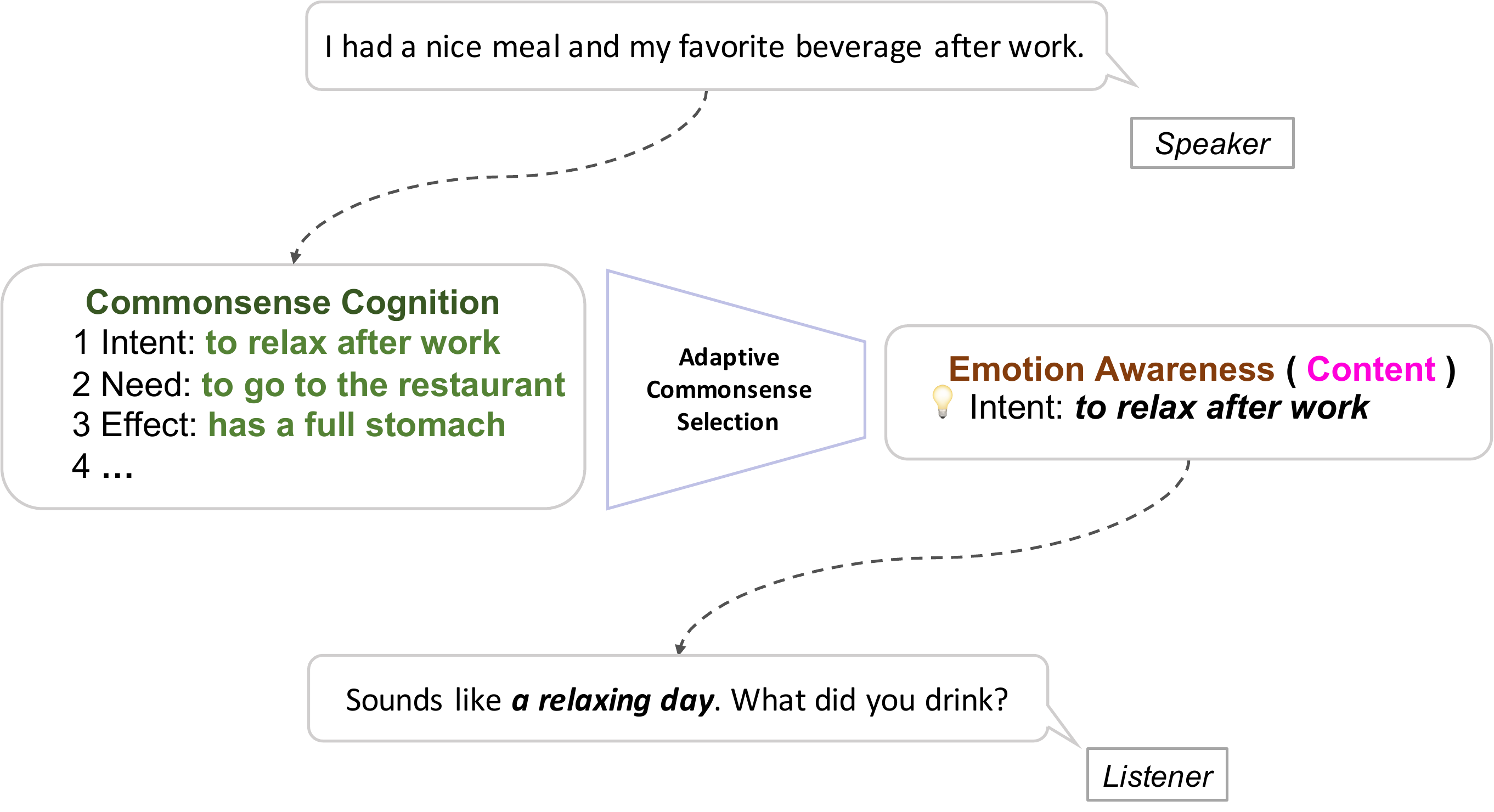}
\caption{The framework of our proposed empathetic dialogue generation. The listener acknowledges speaker's feeling with the adaptive commonsense selection and respond with respect to the emotion status of speaker. }
\label{fig1}
\end{figure}

Yet, understanding speaker's emotion and showing the contextually appropriate comprehension of her/his situation are still challenges in empathetic conversations. When interacting with a dialogue system, the speakers are not expected to explicitly share all the information about their situation and how they may feel. As humans, we use our commonsense knowledge to make connections between what is explicitly mentioned and what is implied. Hence, to address above issues, some prior works \cite{zhou2018commonsense,wu2020diverse} implement external knowledge to identify the speaker's situation, to acknowledge the speaker's status and to bring diversity for generated response.

However, straightforward knowledge merging method confuses the system and the response consistency would be deteriorated. This is demonstrated in Figure \ref{fig1}, where the irrelevant knowledge (\textit{Need}) may potentially form empathetic responses, which conflicts with the information about speaker's emotion (\textit{content}). Accordingly,  the speaker displays the satisfaction of her/his experience, which provides potential informative cognitions based on one unified commonsense. We can assume that if the most appropriate commonsense cognition (\textit{Intent}) is selected with respect to emotion status, the generated response shows better consistency and empathy. Therefore, we believe dialogue systems with rectified knowledge, which aims at unifying the contextual emotion, lead to more consistent and empathetic responses.

In this paper, we address the task of empathetic dialogue generation by dynamically infusing commonsense knowledge. Such additional commonsense knowledge is used to improve the cognitive understanding about the speaker’s situation and feelings, thus enhance the empathy expression in the generated responses. Meanwhile, the dynamical selection stage avoids the confusion of knowledge in dialogue system and enhance the response consistency with context history. In general, our main contributions are summarized as follows:
\begin{itemize}
\item We introduce a novel approach that incorporates the inferred commonsense knowledge to enhance empathetic response generation.
\item We propose an effective knowledge selecting paradigm that could dynamically select the commonsense knowledge, which is most relevant to speaker’s cognitive empathy. To the best of our knowledge, it is the ﬁrst work to study commonsense knowledge dynamical selection for empathetic dialogue generation.
\item Experiments show that with incorporating the selected commonsense, our model is able to generate more empathetic and interpretable responses compared with the previous methods.
\end{itemize}

\section{Related Works}
\subsection{Empathetic Dialogue Generation}
In recent years, research on implementing empathy in open domain dialogue systems and generating empathetic responses has gained considerable attention. Rashkin et al. (\citeyear{Rashkin:2019}) consider a richer and evenly distributed set of emotions and release a dataset EmpatheticDialogues, where a listener responds to a speaker who is under an emotional situation in an empathetic way. Ghosal et al. (\citeyear{ghosal2020cosmic}) demonstrate that detecting the speaker's emotion is an essential part of generating empathetic responses. Prior studies on emotion-related conversational systems mainly focused on rule-based systems, which heavily rely on hand-craft features \cite{zhou2018mojitalk,zhou2018ecm}. Recently, many neural emotional dialogue generation approaches have been explored to control the emotional expression in the target response \cite{Lin:2019,Majumder:2020}. However, Li et al. (\citeyear{Li:2021empgd}) reveal that conventional empathetic conversation systems face an emotional inconsistency problem as they strive to produce emotionally rich responses based on predefined user-input emotions.

\subsection{Connecting Knowledge and Dialogue}
Leveraging knowledge from commonsense knowledge base has been demonstrated for gaining a better understanding of the implied emotions within the context \cite{Tu:2022,lee2022improving}. ConceptNet \cite{speer2017conceptnet} and ATOMIC \cite{sap2019atomic} are commonsense knowledge bases. ConceptNet consists of 36 relations focusing mostly on taxonomic, lexical and physical commonsense knowledge. Distinguished from ConceptNet, ATOMIC consists 9 relations that cover social commonsense knowledge including event-centered causes and effects as well as person-related mental states. Both Zhou et al. (\citeyear{zhou2018commonsense}) and Zhang et al. (\citeyear{Zhang:2019CKG}) introduce knowledge triplets from ConceptNet into open-domain response generation. Recently, Li et al. (\citeyear{Li:2021towards}) and Zhong et al. (\citeyear{zhong2021care}) exploit ConceptNet to enhance emotion reasoning for response generation. Ghosal et al. (\citeyear{ghosal2020cosmic}) utilizes ATOMIC in emotional dialogue modeling for emotion identification. Sabour et al. (\citeyear{Sabour:2021}) leverages commonsense from ATOMIC to improve the understanding of speaker's situations and feelings.

Therefore, enabling dialogue systems to leverage commonsense and driving implications from the speaker's explicit statements are highly beneficial for more empathetic responses. In this work, we focus on the task of empathetic dialogue generation on EmpatheticDialogues dataset, and pay attention to addressing social related commonsense knowledge from ATOMIC. For each event, we use the social relations in ATOMIC to infer the commonsense knowledge about the person involved in the event. We adopt COMET \cite{bosselut-etal-2019-comet} to generate commonsense sentences for the given events. This model is pre-trained on triplets from ATOMIC and then fine tuned on $\textnormal{ATOMIC}^{20}_{20}$ \cite{hwang2021comet}, so that is more suitable for inferring knowledge regarding unseen events in the original ATOMIC daily basis dataset.

\section{Methodology}
\begin{figure}[t]
\centering
\includegraphics[width=0.48\textwidth]{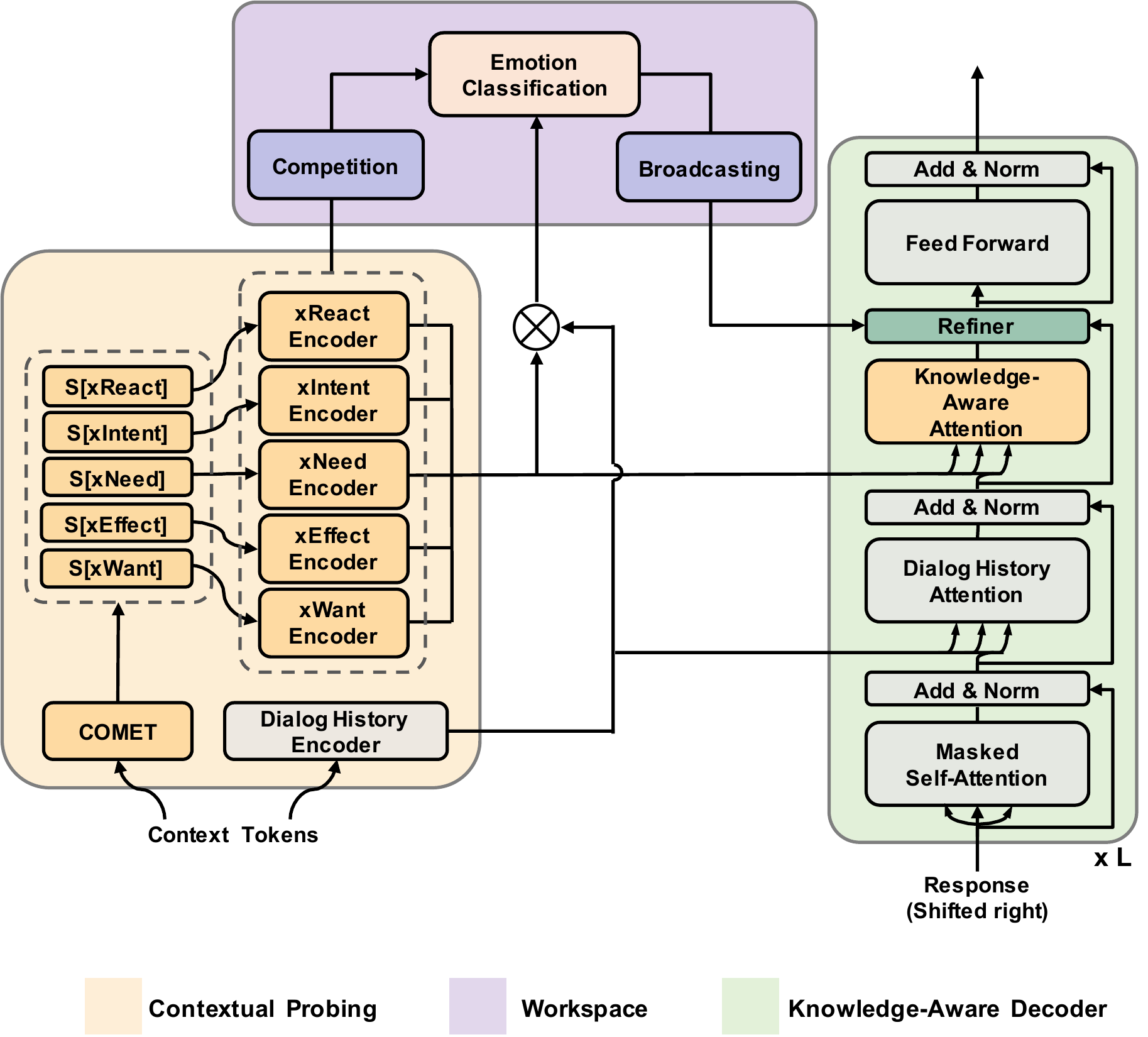}
\caption{The architecture of our framework. It consists of three modules: (1) Contextual Probing enhances dialogue comprehension by commonsense knowledge; (2) Workspace adaptively modifies the cognition of speaker's status; (3) Knowledge-Aware decoder generates empathetic responses.
}
\label{fig2}
\end{figure}

 Our proposed model is built upon the Transformer-based pre-trained
 language model to generate listener’s utterance. Each conversation process of the model is mainly divided into three stages: contextual probing, contextual unification workspace and knowledge-aware decoder. The overview of our model is illustrated in Figure \ref{fig2}. 

\subsection{Task Formulation}
The task requires a dialogue model to play the role of the listener and generate empathetic responses. Formally, let $ U = [u_1,\ u_2,\ ...,\ u_{n-1}]$ denote a dialogue history of $n - 1$ utterances, where $u_i = [w^i_1,\ w^i_2,\ ...,\ w^i_{M_i} ]$ is the i-th utterance that consists of $M_i$ words. Let $K = \{k_i\}$ denote the commonsense knowledge generated from COMET, where $k_i$ is the empathetic commonsense inference knowledge. Our goal is to generate a response Y using historical utterance $U$ and commonsense knowledge $K$ as input. A dialogue history encoder to encode $U$, a knowledge encoder to encoder $K$, and a decoder to incorporate dialog history, dynamically select knowledge and generate response.

\subsection{Contextual Probing}
To obtain semantic representations of the dialog history and the knowledge from ATOMIC, we divide the context probing part into context encoding and knowledge acquisition. 

\subsubsection{Context Encoding}
We concatenate the utterances in the dialogue history and prepend a special token $[CLS]$ to obtain the dialogue historical context input 
$U = [CLS] \oplus u_1 \oplus u_2 \oplus ... \oplus u_{n-1}$, 
where $\oplus$ is the concatenation operation. Then, we use the final hidden representation of $[CLS]$ as the representation of the whole sequence.

We use BART encoder part to acquire the contextual representation. The sequence $U$ is fed into the encoder, and the hidden state of the encoder token:
\begin{equation}
    \mathbf{z}_{\textnormal{ctx}} = \mathbf{Enc}_{\textnormal{ctx}}(U), 
\end{equation}
where $\mathbf{z}_{\textnormal{ctx}} \in \mathbb{R}^{L \times  d}$, $L$ is the length of the sequence, and $d$ is the hidden size of the context encoder.

\subsubsection{Knowledge Acquisition}
In ATOMIC, six relations could be inferred for the person X involved in the event: the effect of the event on X ($xEffect$), X's reaction to the event ($xReact$), X's intent before the event ($xIntent$), what X need in order for the event to happen ($xNeed$), what X would want after the event($xWant$), and an inferred attribute of X’s characteristics ($xAttr$). Since predicting a person’s attributes involves judging the other person, which is not included in the empathetic process, we 
ignore $xAttr$ in our approach and use the remaining five relations.

For input sequence $U$, we respectively append five special relation tokens ($\textnormal{[xReact]}$, $\textnormal{[xWant]}$, $\textnormal{[xNeed]}$, $\textnormal{[xIntent]}$, $\textnormal{[xEffect]}$) to the last utterance in the dialogue history and then use COMET to generate $k$ commonsense inferences $S^r$ =  [$cs^r_1,\ cs^r_2,\ ...\ ,\ cs^r_k$] per relation $r$, where $r \in \{xReact,\ xWant,\ xNeed,\ xIntent,\ xEffect\}$.

For each relation, we concatenate the generated commonsense inferences to obtain its commonsense sequence $CS_r = cs^r_1 \oplus cs^r_2 \oplus ... \oplus cs^r_k$, which demonstrates the knowledge regarding the speaker's dialogue state (i.e. emotion and situation). Accordingly, similar to the previous section, we prepend $[CLS]$ to the sequences denoted as $\mathbf{E}_{CS_{r}}$, which then are fed to five separate commonsense knowledge encoders, as shown in the 
contextual probing part of Figure \ref{fig2}:
\begin{equation}
    \mathbf{Z}_r = \mathbf{Enc}_{Kno}(\mathbf{E}_{CS_{r}} ),
\end{equation}
where $\mathbf{Z}_r \in \mathbb{R}^{l_r \times  d}$, $l_r$ is the lengths of the commonsense inference sequences. 

Then, we utilize the hidden vector of $[CLS]$ as the representation for each relation, and through average operation we obtain the fused representation $\mathbf{z}_r = Average(\mathbf{Z}_r[0]) \in \mathbb{R}^d$ for all relations.

\subsection{Contextual Unification Workspace}
To better leverage the hidden representation from knowledge acquisition and context encoding, we apply the workspace module for unifying contextual information according to emotion label. The workspace consists of two parts: emotion classification for identifying speaker's status, and adaptive knowledge selection for excluding irrelevant knowledge representation. 

\subsubsection{Emotion Classification}
In contrast to concatenating the representations at a sequence level, we use point-wise addition to fuse the additional knowledge in the sequence, i.e., the fusing of knowledge and the context representation:
\begin{equation}
\mathbf{z}_f = \mathbf{z}_r + \mathbf{z}_{\textnormal{ctx}}.  
\end{equation}

In order to acquire a more accurate prediction of the speaker's emotion, given that we are provided with an emotion label $e$ for each conversation, we use the infused representation of knowledge and context representation to perform emotion classification. We also pass $\textbf{z}_f$ through a linear layer $g_{\theta} $, followed by a softmax operation to produce the emotion category distribution $P_{\textnormal{emo}} \in \mathbb{R}^q$, where $q$ is the number of available emotion categories:
\begin{equation}
    P_{\textnormal{emo}} = \mathrm{Softmax}(g_{\theta}({\mathbf{z}_{f})} ),    
\end{equation}
where $\theta \in \mathbb{R}^{d \times  q}$ is the weight vector for the linear layer. During training, we optimize these weights by minimizing the Cross-Entropy (CE) loss between the emotion category distribution $P_{\textnormal{emo}}$ and the ground truth label $e$: 
\begin{equation}
    \mathcal{L}_{\textnormal{emo}} = - \log(P_{\textnormal{emo}}(e)).
\end{equation}

\subsubsection{Adaptive Knowledge Selection}
We present a knowledge selection method that the decoder can adaptively choose the commonsense representations based on the emotion classification results. 
Given the set of knowledge representation  $\mathbf{Z} = \{ \mathbf{Z}_r[0] \}$, the goal is to choose the most appropriate knowledge relations that satisfy the consistency with the context representation vector $\mathbf{z}_{\textnormal{ctx}}$. By this selection paradigm, the irrelevant relations, which would potentially confused the generated response, will be eliminated, so as to boost the performance of dialogue system. 

Inspired by Global Workspace Theory in cognitive science \cite{ Blum,baars1993cognitive} , the process of  contextual coordination is realized by eliminating irrelevant cognition. We therefore implement the label of emotion as the coordination of context and the $\mathcal{L_\textnormal{emo}}(g_{\theta}(\mathbf{z}),g_{\theta}(\mathbf{z}_{\textnormal{ctx}}))$ from the supervised evaluation to eliminate irrelevant cognition. The knowledge selection mechanism is divided into two stages, \textit{competition} and \textit{broadcasting}:  

\begin{itemize}

\item During the \textit{competition} stage, we recursively exclude the irrelevant information of knowledge representation based on the emotion status. Specifically, at iteration $m$, we choose the $\max_{ \mathbf{z} \in \mathbf{Z}}\{\mathcal{L_\textnormal{emo}}(g_{\theta}(\mathbf{z}),g_{\theta}(\mathbf{z}_{\textnormal{ctx}})) \}$ as the most irrelevant knowledge representation. In order to model the influence of knowledge exclusion, we leverage nonlinear regression method \cite{qing:2019,10027603} to calculate the dynamics ${\mathbf{G}} = \nabla_{{\bm{\theta}}} \mathbf{f} \in \mathbb R ^{d \times q}$ of the aforementioned max loss. Please refer to the Appendix for the technical details. After the last iteration, the remaining knowledge representation,  as the winner of competition, is applied for acknowledging the unified speaker's emotion status.

\item  In the \textit{broadcasting} stage, the winner of competition stage will be applied for unifying the combined representation in decoder. Specifically, we realize this stage by adding the dynamics of the selecting process to rectify the knowledge representation.  Thus,  the generated response will less affected by the unrelated information from knowledge encoder in contextual probing module.  

\end{itemize}

We provide Algorithm \ref{alg:algorithm} in Appendix to show the exclusion method. Figure \ref{fig3} displays how the workspace process refine the knowledge representation.



\begin{figure}[t]
\centering
\includegraphics[width=0.43\textwidth]{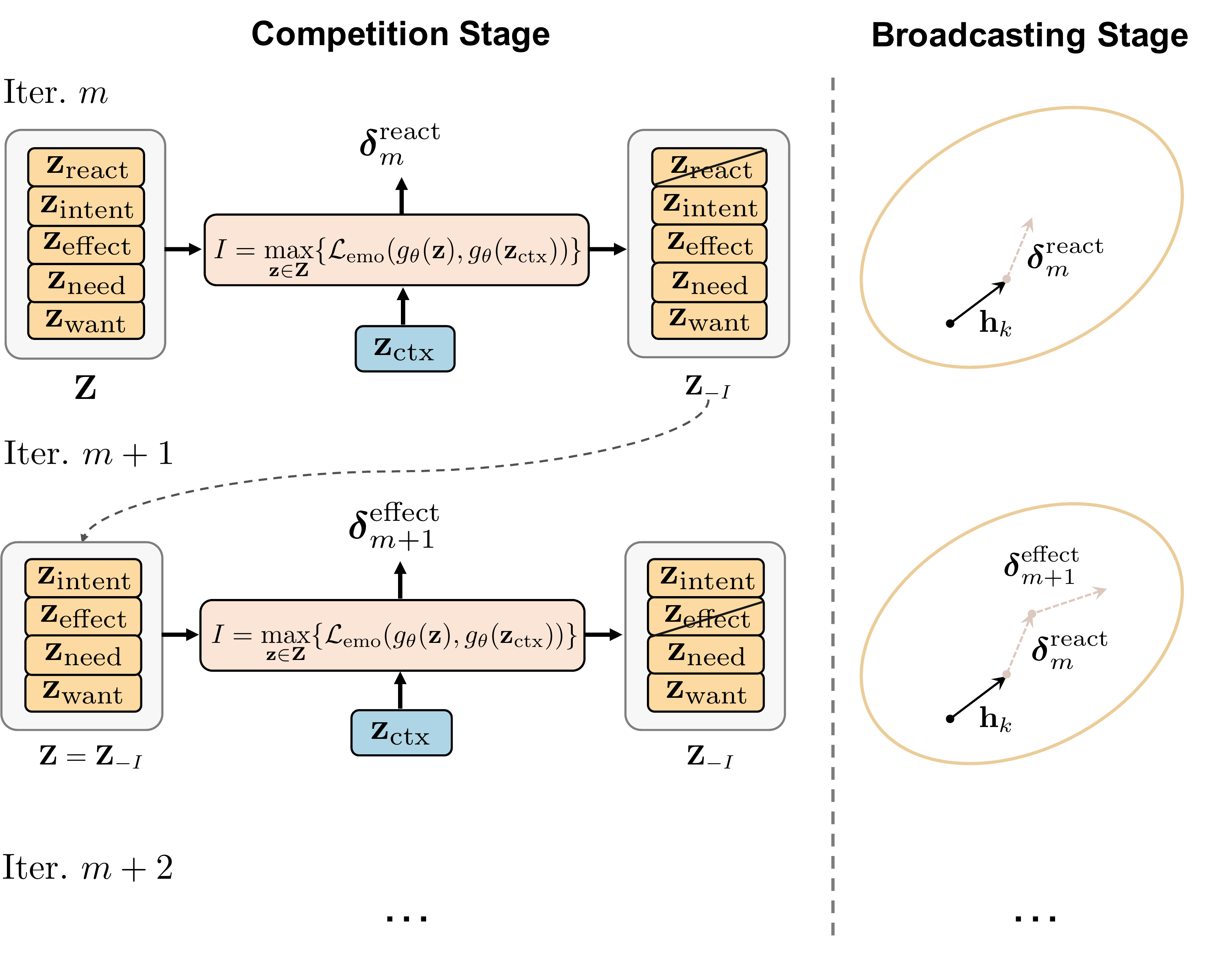}
\caption{The illustration of the workspace mechanism. During competition stage at each iteration, the most irrelevant knowledge, for example the `react', is deleted from the set of knowledge representation, demonstrated by $\mathbf{Z}_{-I} $. The dynamics of the deletion is $\boldsymbol{\delta}$. During broadcasting stage, the knowledge-aware representation $\mathbf{h}_k$ is refined by the dynamic  $\boldsymbol{\delta}$.  }
\label{fig3}
\end{figure}

\subsection{Knowledge-Aware Decoder}
Generally, not all knowledge contributes to the generation of the response, so the model should have the ability to select knowledge. Instead of performing knowledge selection in the encoding phase, we leave it to the decoding phase. As shown in the right part of Figure \ref{fig2}, a knowledge-aware cross attention block is introduced to select knowledge dynamically. Feed the selected knowledge to the context-knowledge refiner, which assists in response generation. The fused knowledge is taken as the input of this block, and then the output of this block is refined to exploit the knowledge contributions.  

\subsubsection{Knowledge Refiner}
In order to refine the context and knowledge contributions in each layer, we replace the residual addition to a refine gate after the knowledge-aware attention block. Denote $\mathbf{h}_k$ as output of knowledge-aware attention block and $\mathbf{h}_c$ as the residual from the previous block, the output of refiner can be expressed by:
\begin{equation}
    R_f(\widetilde{\mathbf{h}}_k, \mathbf{h}_c) = \alpha \cdot \mathbf{LN}(\widetilde{\mathbf{h}}_k) + (1-\alpha) \cdot  \mathbf{h}_c
\end{equation}
\begin{equation}
 \widetilde{\mathbf{h}}_{k} = \mathbf{h}_{k} + \bm{\delta}_m 
\end{equation}
\begin{equation}
    \alpha =\sigma (\mathbf{w} \cdot [\widetilde{\mathbf{h}}_k; \mathbf{h}_c])   
\end{equation}
Where $\mathbf{LN}$ is a linear layer, $\widetilde{\mathbf{h}}_k$ is the rectified knowledge representation, $ \mathbf{w} \in \mathbb{R}^{2d}$ is a learnable parameter and $\sigma$ denotes sigmoid function.

\subsubsection{Response Generation}
Lastly, the target response $Y = [y_1,\ y_2,\ . . . ,\ y_T]$ with length $T$, which is generated by the decoder token by token by using the embeddings of the tokens that have been generated and the commonsense-refined contextual representation $R_f(\widetilde{\mathbf{h}}_k, \mathbf{h}_c)$, which has fused the information from both the context and the commonsense inferences. We adopt the standard negative log-likelihood ($NLL$) loss on the target response $Y$:
\begin{equation}
    \mathcal{L}_{\textnormal{nll}} = - \sum ^T_{t=1}\log(y|(\mathbf{U}, \mathbf{K}), y_{<t}).
\end{equation}

\subsection{Training Objectives}
All the parameters for our proposed model are trained and optimized based on the weighted sum of the two mentioned losses:
\begin{equation}
   \mathcal{L} = \mathcal{L}_{\textnormal{nll}}  + \gamma \mathcal{L}_{\textnormal{emo}},
\end{equation}
where $\gamma$ is hyper-parameter that we use to control the influence of the these losses. In our experiments, we set $\gamma = 1$.




\section{Experiments}

\begin{table*}[t]
\fontsize{9pt}{9pt}\selectfont
\centering
\begin{tabular}{@{}lcccccccccc@{}}
\toprule
\multicolumn{1}{c}{\textbf{Models}} & \textbf{PPL}   & \textbf{B-1} & \textbf{B-2} & \textbf{B-3} & \textbf{B-4} & \textbf{R-1} & \textbf{R-2} & \textbf{Dist-1} & \textbf{Dist-2} & \textbf{Acc}   \\ \midrule
Transformer     & 37.62          & 18.07           & 8.34            & 4.57            & 2.86            & 17.22            & 4.21             & 0.36            & 1.35            & --             \\
Multi-TRS       & 37.50           & 18.78           & 8.55            & 4.70             & 2.95            & 16.85            & 4.21             & 0.35            & 1.27            & 33.95          \\
MoEL            & 36.60           & 18.07           & 8.30             & 4.37            & 2.65            & 18.24            & 4.81             & 0.59            & 2.64            & 31.74          \\
MIME            & 37.24          & 18.60            & 8.39            & 4.54            & 2.81            & 17.08            & 4.05             & 0.47            & 1.66            & 30.96          \\
EmpDG           & 37.43          & 19.96           & 9.11            & 4.74            & 2.80             & 18.02            & 4.43             & 0.46            & 1.99            & 31.65          \\
CEM             & 36.33          & 16.12           & 7.29            & 4.06            & 2.03            & 15.77            & 4.50              & 0.62            & 2.39            & 36.84          \\ \midrule
Ours            & 16.08          & \textbf{21.73}  & \textbf{10.62}  & \textbf{6.24}   & \textbf{4.09}   & \textbf{19.77}   & \textbf{5.65}    & \textbf{2.19}   & \textbf{9.61}   & \textbf{49.16} \\
w/o A$\ast$         & 15.41 & 19.50            & 9.54            & 5.52            & 3.62            & 19.35            & 5.57             & 2.16            & 8.87            & 46.47          \\
w/o Knowledge   & \textbf{15.24}          & 20.11           & 9.86            & 5.72            & 3.73            & 19.72            & 5.82             & 2.08            & 8.59            & 44.87          \\
w/o Context     & 15.62          & 20.45           & 9.98            & 5.78            & 3.74            & 19.88            & 5.78             & 1.82            & 7.41            & 46.34          \\
\bottomrule
\end{tabular}
\caption{Results of automatic evaluation. A$\ast$ represents the adaptive knowledge selection method in the workspace module. }
\label{tb:autoeval}
\end{table*}

\subsection{Datasets}
We conduct our experiments on the EmpatheticDialogues, a large-scale multi-turn dataset containing 25k empathetic conversations between crowd sourcing workers. The dataset also provides an emotion label for each conversation from the total 32 available emotions.

\subsection{Baselines}
We select the following baseline models for comparison on EmpatheticDialogues:  (1) \textbf{Transformer} \cite{NIPS2017_3f5ee243}: An original Transformer, which is trained to optimize the NLL loss. (2) \textbf{Multi-TRS} \cite{Rashkin:2019}: A variation of the Transformer for multitask that trained to jointly optimize an additional cross-entropy loss for emotion classification with the NLL loss. (3) \textbf{MoEL} \cite{Lin:2019}: A Transformer-based model that uses 32 emotion-specific decoders to generate a response. Therefore, each decoder is optimized to respond appropriately for each emotion. (4) \textbf{MIME} \cite{Majumder:2020}: Another Transformer-based model that mimics the context emotion to a varying degree considering its negative and positive emotions, and then generates empathetic response based on the blend of these two emotions. (5) \textbf{EmpDG} \cite{Li:2021empgd}: A multi-resolution adversarial framework which applies an empathetic generator to produce empathetic responses and an interactive discriminator to ensure that the generated responses are consistent with the context and are also empathetic. (6) \textbf{CEM} \cite{Sabour:2021}: An empathetic generation approach which leverages commonsense to draw more information about the speaker's situation and uses this additional information to further enhance the empathy expression in generated responses. 

\subsection{Implementation Details}
We implement all the models using PyTorch and use the encoder and decoder from base version of BART in our work. We use Adam optimizer with initial learning rate 0.00005 in 5 epochs. The batch size is 16. The max sequence length in source and target is 256 and 64 respectively. We use the same 8:1:1 train/valid/test split as provided by Rashkin et al. (\citeyear{Rashkin:2019}). In each experiment, we apply an early stop mechanism to prevent the model from over fitting, and then report the test results of the optimal model on the test set. All our training and test results were performed on 32GB Tesla V100 GPU.

\subsection{Evaluation Metrics}

\subsubsection{Automatic Evaluation}
We employ Perplexity (PPL), corpus-level BLEU (B-n), sentence-level ROUGE (R-n) and Distinct-n (Dist-n) as our main automatic metrics. Perplexity represents the model's confidence in its set of candidate responses, with higher confidence resulting in a lower PPL. This can be used to evaluate the general quality of the generated responses. Response with higher BLEU and ROUGE is closer to the ground-truth. Distinct-n measures the proportion of unique n-grams in the generated responses and is commonly used to evaluate generation diversity. In addition, since our proposed model and most baseline models perform emotion classification as part of their training process, we also report the prediction accuracy (Acc). 

\subsubsection{Human Evaluation}
Following the methods in CEM, we conduct an aspect-based pairwise preference test. That is, for a given context, we pair our model’s response with a response from the baselines and ask annotators to give each response a rating score from four aspects: 1) Coherence (\textbf{Coh.}): which response is more coherent in content and relevant to the context; 2) Empathy (\textbf{Emp.}): which response shows more understanding of the speaker's situation and presents a more appropriate emotion; 3) Informativeness (\textbf{Inf.}): which response conveys more information about the context. 4) Continuity (\textbf{Con.}): which response ignites the speaker's more desire to continue the conversation. Then, we randomly sample 100 response pairs and totally shuffle the response order in each sample. We assign crowd sourcing workers to annotate each pair on a scale of 1 to 5.

\subsection{Evaluation Results}
\subsubsection{Automatic Evaluation Results}
Table \ref{tb:autoeval} reports the evaluation results on automatic metrics. Ours model achieves the lowest perplexity, which suggests the overall quality of our generated responses is higher than the baselines, approximately 56\% lower than CEM. In addition, our model also considerably outperforms the baselines in terms of Dist-n, BLEU-n and ROUGE-n, which highlights the diversity of the responses and the relevance between generated response and speaker's situation. In terms of emotion classification, our model had a much higher accuracy compared to the baselines, nearly 34\% higher than CEM, which suggests the adaptive selection of commonsense knowledge is pivotal for detecting the speaker's emotion.

Table \ref{tb:lowresource} reports the evaluation results on low-resource training set, and we have the following observations: (1) In the full-data scenario, our model achieves start-of-the-art performance by infusing commonsense knowledge, which means that the importance of knowledge in dialogue generation. Besides, reducing the number of training samples has effect on model performance, but not that much, for that even the model using 1/4 data still has the approximate values in PPL, BLEU-n, ROUGE-n and Dist-n compared with the model using full data. (2) In the 1/8 training data scenario, our model achieves the comparable performance with baselines even though them leveraged all training data. (3) Responses generated by our model have higher Dist-n in low-resources scenarios, which means that our model can better obtain information from multiple knowledge and generate more diverse texts.

\subsubsection{Ablation Studies}
We conduct ablation studies to verify the effectiveness of each of the components in emotion classification and the generation performance. Specifically, we design three variants: \textit{workspace}, \textit{knowledge} and \textit{context}. It is worth noting that since \textit{workspace} depends on \textit{knowledge} and \textit{context}, when \textit{knowledge} or \textit{context} module is removed, \textit{workspace} is removed by default: 

\begin{enumerate}
\item w/o Adapter: the mechanism in workspace that used for adaptive commonsense knowledge selection is removed, and the emotion classification is based on none selected commonsense representation;
\item w/o Knowledge: the commonsense knowledge representation used for emotion classification is removed (Equation 6), and the hidden representation of the [CLS] token from the encoded context is used for emotion classification;
\item w/o Context: the context representation used for emotion classification is neglected (Equation 6), but keep the affective and cognitive commonsense knowledge representations;
\end{enumerate}

\begin{table}[]
\fontsize{9pt}{9pt}\selectfont
\begin{tabular}{@{}lcccccc@{}}
\toprule
\multicolumn{1}{c}{\textbf{Models}} & \textbf{PPL} & \textbf{B-2} & \textbf{B-4} & \textbf{R-1} & \textbf{R-2} & \textbf{Acc} \\ \midrule
Ours                                & 16.08        & 10.62           & 4.09            & 19.77            & 5.65             & 49.16        \\
1/2 Data                       & 16.57        & 10.00              & 3.58            & 19.56            & 5.35             & 40.00           \\
1/4 Data                       & 16.43        & 9.72            & 3.32            & 18.61            & 4.83             & 34.24        \\
1/8 Data                       & 18.51        & 9.33            & 3.29            & 18.61            & 4.82             & 33.80         \\
1/16 Data                       & 44.71        & 8.77            & 2.56            & 17.04            & 4.06             & 25.29         \\
Zero Data                      & 100+         & 3.99            & 0.86            & 10.20             & 1.09             & 2.60          \\ \bottomrule
\end{tabular}
\caption{Evaluation results on low-resource training set of EmpatheticDialogues. }
\label{tb:lowresource}
\end{table}

The obtained results are shown in Table \ref{tb:autoeval}. We observe that reducing the workspace module results in lower classification accuracy as the same as BLEU-n and ROUGE-n. And removing the commonsense knowledge information also impacts the emotion classification accuracy. The above phenomena suggest that information about both the speaker's emotion and their situation are necessary for correctly identifying their feelings, and dynamical knowledge selection is leveraging the knowledge contribution to the cognition response. Removing those components leads to lower Dist-n scores but higher perplexity, which indicates the effectiveness of those components in generating more diverse responses.

\begin{table*}[t]
\centering
\fontsize{10pt}{10pt}\selectfont
\begin{tabular}{@{}l|l@{}}
\toprule
\textbf{Context }        & I ended up \textbf{\textit{staying at home}} for the holidays.                                                                                                                                                                                                                                                                                                                                                         \\ \midrule
\textbf{MoEL  }             & Where did you go?                                                                                                                                                                                                                                                                                                                                                                                   \\
\textbf{MIME   }            & I am sorry to hear that. What happened?                                                                                                                                                                                                                                                                                                                                                            \\
\textbf{EmpDG  }          & Oh no! What happened?                                                                                                                                                                                                                                                                                                                                                                              \\
\textbf{CEM}                & Did you enjoy it?                                                                                                                                                                                                                                                                                                                                                                                    \\ \midrule
\textbf{Selection Process} & \begin{tabular}[c]{@{}l@{}}xEffect $ \rightarrow$ xReact $ \rightarrow$ xWant $ \rightarrow$  xNeed $ \rightarrow$ xIntend \end{tabular}   \\

\textbf{Selected Knowledge} & xIntent: To be alone; \textbf{\textit{To be home}}; To relax; To rest                                                                                                                                                                                                                                                                                                                                         \\
\textbf{Emotion}            & \textbf{\textit{Lonely}}                                                                                                                                                                                                                                                                                                                                                                                               \\
\textbf{Ours    }           & Oh, I am sorry to hear that. \colorbox{yellow}{What did you do while staying at home}?                                                                                                                                                                                                                                                                                                                                  \\ \midrule
\textbf{Gold}               & Why is that? Comfort?                                                                                                                                                                                                                                                                                                                                                                              \\ \midrule \midrule
\textbf{Context}            & My friend borrowed quite a lot of money from me. I really \textbf{\textit{do believe he will repay me}}.                                                                                                                                                                                                                                                                                                              \\ \midrule
\textbf{MoEL  }             & That is so nice of him. Did you find a good friend?                                                                                                                                                                                                                                                                                                                                                \\
\textbf{MIME  }             & That is a good thing to do.                                                                                                                                                                                                                                                                                                                                                                         \\
\textbf{EmpDG }             & That is a good friend.                                                                                                                                                                                                                                                                                                                                                                              \\
\textbf{CEM }               & That is nice of him.                                                                                                                                                                                                                                                       \\ \midrule
\textbf{Selection Process} & \begin{tabular}[c]{@{}l@{}}xReact $ \rightarrow$ xIntend $ \rightarrow$ xEffect $ \rightarrow$  xWant $ \rightarrow$ xNeed \end{tabular}                                                                                                                   \\ 
\textbf{Selected Knowledge} & xNeed: To ask for a loan; To get a loan; \textbf{ \textit{To ask him to repay}}; To ask for money                                                                                                                                                                                                                                                                                                 \\
\textbf{Emotion}            & \textbf{\textit{Trusting}}                                                                                                                                                                                                                                                                                                                                                                                             \\
\textbf{Ours}               & I am sure \colorbox{yellow}{he will repay you}.                                                                                                                                                                                                                                                                                                                                                                         \\ \midrule
\textbf{Gold}               & You do? That's good, friends can be terrible people to lend too.                                                                                                                                                                                                                                                                                                                                 \\ 
\bottomrule
\end{tabular}
\caption{ We report the case study of generated responses from EmpatheticDiaglogues. The responses with yellow background color demonstrate the awareness to the emotion and the selected knowledge.  }
\label{tb:case}
\end{table*}

\begin{table}[t]
\centering
\fontsize{10pt}{10pt}\selectfont
\begin{tabular}{@{}ccccc@{}}
\toprule
\textbf{Models} & \textbf{Coh.} & \textbf{Emp.} & \textbf{Inf.} & \textbf{Cont.} \\ \midrule
MoEL            & 3.57          & 3.26          & 3.11          & 3.09           \\
MIME            & 3.61          & 3.30          & 3.09          & 3.13           \\
EmpDG           & 3.42          & 3.10          & 2.94          & 2.89           \\
CEM             & 3.90          & 3.49          & 3.08          & 3.19           \\
\midrule
Ours            & \textbf{4.39}          & \textbf{4.13  }        & \textbf{4.18 }         & \textbf{4.24}           \\ \bottomrule
\end{tabular}
\caption{Results of human evaluation. We report the average scores of four aspects. Fleiss kappa of the results is 0.36, which constitutes a fair level of agreement.}
\label{tb:humaneval}
\end{table}

\begin{figure}[t]
\centering
\includegraphics[width=0.47\textwidth]{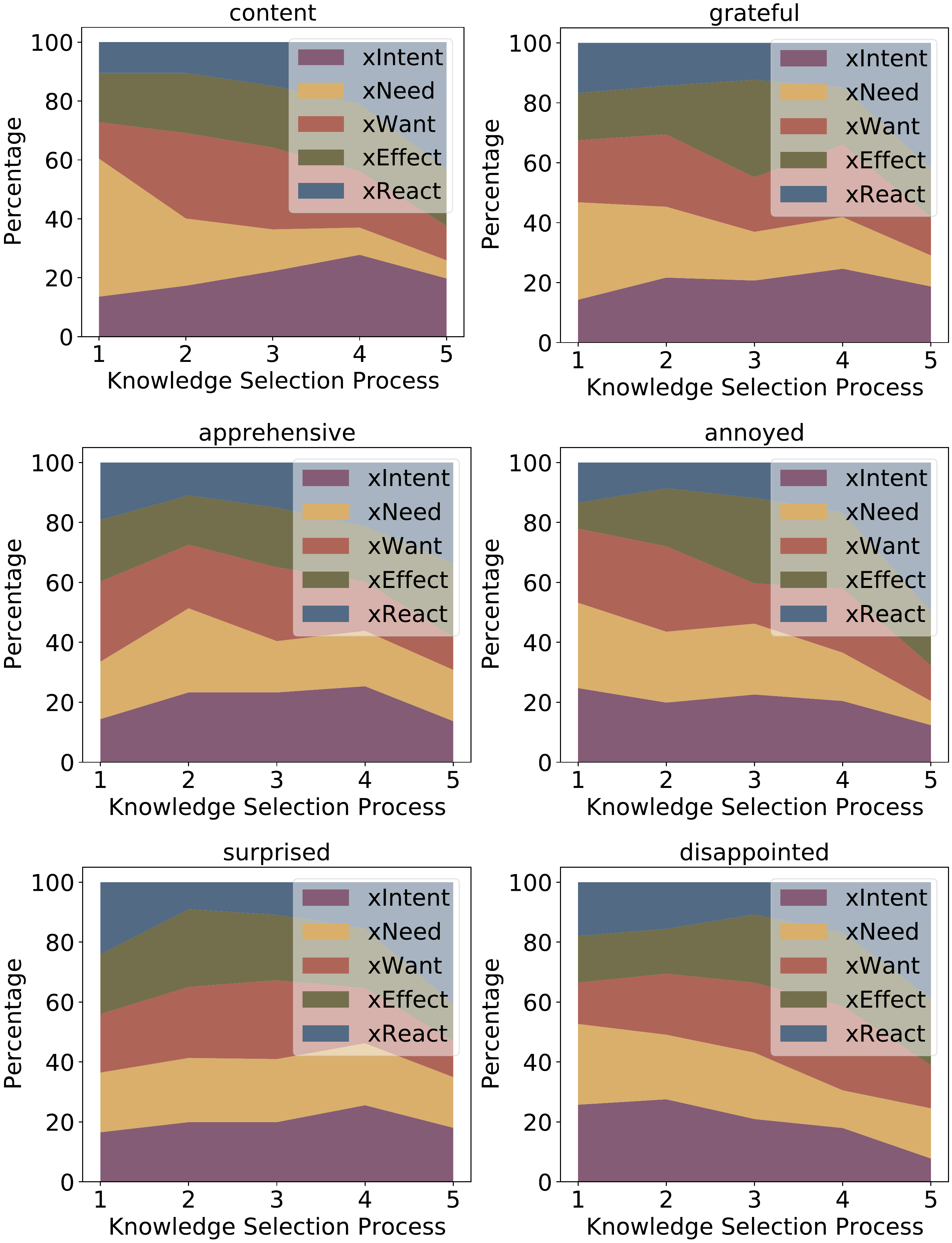}
\caption{Stackplot of the knowledge selection process.}
\label{fig4}
\end{figure}

\subsubsection{Human Evaluation Results}
Table \ref{tb:humaneval} reports the evaluation results on human ratings. We observe that responses from our model are more contextually coherent than those from baselines. Besides, with the enhancement of commonsense knowledge, the response from our model are able to convey more specific and informative content. It is worth to note that, for the aspect of continuity, our model significantly outperforms all the baselines, which suggests that the generated responses may increase speaker's engagement, thus a more intimate emotional expression.

\subsection{Qualitative Studies}
\paragraph{Case Study} Table \ref{tb:case} shows the cases from EmpatheticDialogues, from which we can see that the response of our method outperforms the baselines. We analyze these cases with respect to the four factors evaluated by human. In aspect of \textit{Coherence} and \textit{Informativeness}, our response is more coherent in content and consistent to the context information. For instance, in case one, by the awareness of selected knowledge  `To be home', our method mentions this phrase in response so that the response better acknowledges speaker's intention. However, other methods fail to generate consistent response. It can be observed that MoEL and CEM dismiss the implication that the speaker is alone at home. The workspace module improves \textit{Empathy} and \textit{Continuity}   by selecting the most influential commonsense with respect to the context. In both cases, the selected knowledge corresponds to the speaker's situation, which produces a more meaningful response by showing careness for speakers. 

\paragraph{Efficacy of Knowledge Selection } Selection process illustrates that the most irrelevant knowledge is
selected and eliminated at each iteration. By combining dynamics from the selection process in refiner, the generated sentence gradually focuses on speaker's emotion status, so that our method provides more interpretable knowledge selection process for the dialogue system.
Figure \ref{fig4} provides characteristic of knowledge selection process. It indicates that  workspace module tends to  select inferred knowledge from the relation xReact. Since xReact reflects speaker's reaction to context, our adaptive selection method  potentially provides the consistency between  context and knowledge. 

\section{Conclusions}

In this paper, we improve empathetic dialogue generation by infusing dynamical commonsense knowledge to promote the understanding of the speaker's situation and feelings, which leads to more consistent and empathetic responses. The automatic and human evaluation demonstrate that the effectiveness of our approach in high-quality empathetic response generation. 

\section*{Limitations}
One limitation in this work is the metrics employed in the automatic evaluation. The metrics mainly focus on the quality of generated response and the accuracy of emotion recognition, while automatic evaluation lacks a comprehensive method to evaluate empathy. Another limitation comes from the utilization of the dataset designed for open-domain dialogue system, so that the generated response from the proposed framework is not task-oriented. In the future, we will build empathetic dialogue generation datasets with diverse and task-oriented response, and develop metrics to evaluate the understanding of the speaker’s situation.

\section*{Ethics Statement}
The human evaluation is conducted by the employed  workers, who does not involve privacy issues. We use public datasets to conduct our experiments. Existing packages involved in this work are displayed in the appendix. 

\section*{Acknowledgment}
This work was supported by UniDT's Cognitive Computing and Few Shot Learning Project.


\bibliography{anthology,custom}
\bibliographystyle{acl_natbib}

\newpage
\appendix

\section{The Details of Cognition Dynamics}
\label{sec:appendix}

Our goal is to calculate the effect of knowledge representation on the predictions of the linear transformation function $g_{\bm{\theta}}$ in the \textit{workspace} module. 
The influence of excluding irrelevant knowledge representation can be interpreted as the change of  $\bm{\theta} $ with respect to $\mathcal{L_\textnormal{emo}}$, which is $ \nabla_{\bm{\theta}} \mathcal{L_\textnormal{emo}}(g_{\theta}(\mathbf{z}),g_{\theta}(\mathbf{z}_{\textnormal{ctx}})), \mathbf{z} \in \mathbf{Z} $. Here, $\mathbf{Z} = \{ \mathbf{Z}_r[0] \},  \mathbf{Z}_r[0] \in \mathbb{R}^d$. In order to eliminate the most irrelevant knowledge representation, we  take the $\max(\cdot)$ on loss function with respect to  $\mathbf{z} \in \mathbf{Z}$. However, it is challenging to calculate the gradient when we implement $\max(\cdot)$ on the groups of loss functions, because the above function is non-differentiable. Thus,   
we first bring differentiability for $\max_{ \mathbf{z} \in \mathbf{Z}}\{\mathcal{L_\textnormal{emo}}(g_{\theta}(\mathbf{z}),g_{\theta}(\mathbf{z}_{\textnormal{ctx}})) \} $. To simplify notation, objective function is set as $\Phi({\bm{\theta}})=\max_{1\leq j \leq J}f_j({\bm{\theta}}),\ f_j({\bm{\theta}})=\mathcal{L_\textnormal{emo}}(g_{\theta}(\mathbf{z}_j),g_{\theta}(\mathbf{z}_{\textnormal{ctx}})) .
$ Here, $f_j$ denotes the loss function with respect to knowledge representation $ \mathbf{z}_j $  and each $f_j$ is differentiable. $g_{\bm{\theta}}$ is the parametric linear layer. Then, calculating the gradient of $ \bm{\theta} $ turns into the following discrete mini-max problem:
\begin{equation}
\min_{{\bm{\theta}} \in\mathbb{R}^d}\max_{1\leq j \leq J} f_j({\bm{\theta}}).
\end{equation}
In order to smooth objective function $ \Phi $ during the iteration $m$, we linearize $f_j$ at ${\bm{\theta}}_m$ and obtain the convex approximation of $\Phi $ as
\begin{equation}
\label{eq:hat_phi}
\hat\Phi({\bm{\theta}}) =  \max_{1 \leq j \leq J}\{ \underbrace{ f_j({\bm{\theta}}_m) + \langle \nabla f_j({\bm{\theta}}_m),{\bm{\theta}} - {\bm{\theta}}_m \rangle \} }_{\rm{linearization \ term}}.
\end{equation}
The linearization term smooths $\max(\cdot)$ function. Next step is to find descent direction, which minimizes $\hat{\Phi} $. However, $\hat{\Phi} $ is not strictly convex with respect to $ {\bm{\theta}}$, the algorithm may not reach global minimum. So a regularization term $  \| {\bm{\theta}} - {\bm{\theta}}_m \|_2 $ is added for finding stable descent direction. Denote the descent direction  ${\bm{\delta}} = {\bm{\theta}} - {\bm{\theta}}_m$, the discrete mini-max problem now is equivalent to
\begin{subequations}
\label{eq:norm}
\begin{align}
\min_{{\bm{\delta}},\nu} & \quad \left \| {\bm{\delta}} \|_2 +\nu\right. \\
\textrm{s.t.} \quad  f_j({\bm{\theta}}_m) + \langle \nabla f_j(&{\bm{\theta}}_m),{\bm{\delta}}\rangle \leq \nu, \ \forall \space 1\leq j \leq J.
\end{align}
\end{subequations}
Problem \eqref{eq:norm} is a semi-definite quadratic programming (QP) problem since we choose $\ell_{2}$ norm as the regularization term. When the number of datapoints in subgroup is large, widely-used QP algorithms, such as active-set method, are time-consuming. Thus we turn to the dual problem.
Consider the Lagrange multiplier for problem \eqref{eq:norm},
\begin{align}
L({\bm{\delta}},\nu;&{\bm{\lambda}}) = \frac 12 \|{\bm{\delta}}\|^2+\nu \nonumber \\
&+ \sum^{J}_{j=1}\lambda_j(f_j({\bm{\theta}}_m)+\langle\nabla f_j({\bm{\theta}}_m),{\bm{\delta}} \rangle-\nu).
\end{align}
By strong duality theorem, the minimum of original problem is equal to the maximum of dual problem under specific constrains:
\begin{align}
\min_{{\bm{\delta}},\nu} \max_{{\bm{\lambda}}\ge0} L({\bm{\delta}},\nu;{\bm{\lambda}}) =\max_{{\bm{\lambda}}\ge0}\min_{{\bm{\delta}},\nu}L({\bm{\delta}},\nu;{\bm{\lambda}})
\end{align}
Let $\mathbf{f} = (f_1,\cdot \cdot \cdot ,f_J)^T$ and ${\mathbf{G}} = \nabla_{{\bm{\theta}}} \mathbf{f}\in \mathbb R ^{d \times q} $. By setting $\mathbf{e} = \mathbf{1}$, the above problem is equivalent to
\begin{align}
\max_{{\bm{\lambda}}\ge0}\min_{{\bm{\delta}},\nu} \big( \frac12 \| {\bm{\delta}} \|^2 + \nu+{\bm{\lambda}}^T(\mathbf{f}  +{\mathbf{G}}{\bm{\delta}} - \nu \mathbf{e})\big) .
\end{align}
Note that
\begin{align}
    \frac12 \|{\bm{\delta}}\|^2 &+ \nu + {\bm{\lambda}}^T(\mathbf{f}  +{\mathbf{G}}{\bm{\delta}}- \nu \mathbf{e}) \nonumber \\
    &=\frac12 \| {\bm{\delta}} \|^2 +{\bm{\lambda}}^T(\mathbf{f} + {\mathbf{G}}{\bm{\delta}})+\nu(1 - {\bm{\lambda}}^T\mathbf{e}).
\end{align}
If $1 - {\bm{\lambda}}^T\mathbf{e} \neq 0$, the objective function will be $-\infty$. Thus, we must have $1 - {\bm{\lambda}}^T\mathbf{e} = 0 $ when the maximum is attained. The problem is converted to
\begin{align}
\label{eq:16}
\max_{{\lambda}_i\ge0,\sum^{J}_{i=1}\lambda_i=1} \min_{\bm{\delta}} \frac 12  \| {\bm{\delta}} \|^2 +{\bm{\lambda}}^T {\mathbf{G}}{\bm{\delta}} +{\bm{\lambda}}^T \mathbf{f}  .
\end{align}
Let the gradient of the inner minimization term to be zero, we have solution $ {\bm{\delta}} = -{\mathbf{G}}^T {\bm{\lambda}}$. By changing the sign of \eqref{eq:16}, the maximization term is reduced to
\begin{subequations}
\begin{align}
\min_{\bm{\lambda}} &\quad (\frac12 {\bm{\lambda}} ^T {\mathbf{G}}{\mathbf{G}}^T {\bm{\lambda}} - {\bm{\lambda}}^T\mathbf{f})  \\
\textrm{s.t.} &\quad \sum^{J}_{i=1}\lambda_i =1,\lambda_i\geq 0.
\end{align}
\end{subequations}
Suppose ${\bm{\lambda}}$ is the solution of the QP problem \eqref{eq:norm}, then  $ {\bm{\delta}} = - {\mathbf{G}}^T {\bm{\lambda}} $  is the solution of problem above. Thus, we have the ${\bm{\delta}} $ as the change of eliminating irrelevant knowledge representation $ \mathbf{z}$. By adding ${\bm{\delta}} $ to the refiner in decoder module, the final generated response would be less affected by the irrelevant knowledge. 
The effect of ${\bm{\delta}} $ is demonstrated by the generated responses  in Table \ref{tb:caseous}, and we also display how the elimination of irrelevant knowledge boost the performance.

\section{Involved Existing Packages}
Existing packages involved in this work include: 1) the open source codes, models weights and generated outcomes of Transformer \citep{NIPS2017_3f5ee243}, Multi-TRS \citep{Rashkin:2019}, MoEL \citep{Lin:2019}, MIME \citep{Majumder:2021}, EmpDG \citep{Li:2021empgd}, CEM \citep{Sabour:2021}, and 2) the evaluation metrics   from Natural Language Toolkit \cite{bird2009natural}.

\section{Additional Case Study }
\begin{figure*}[t]
\centering
\includegraphics[width=1\textwidth]{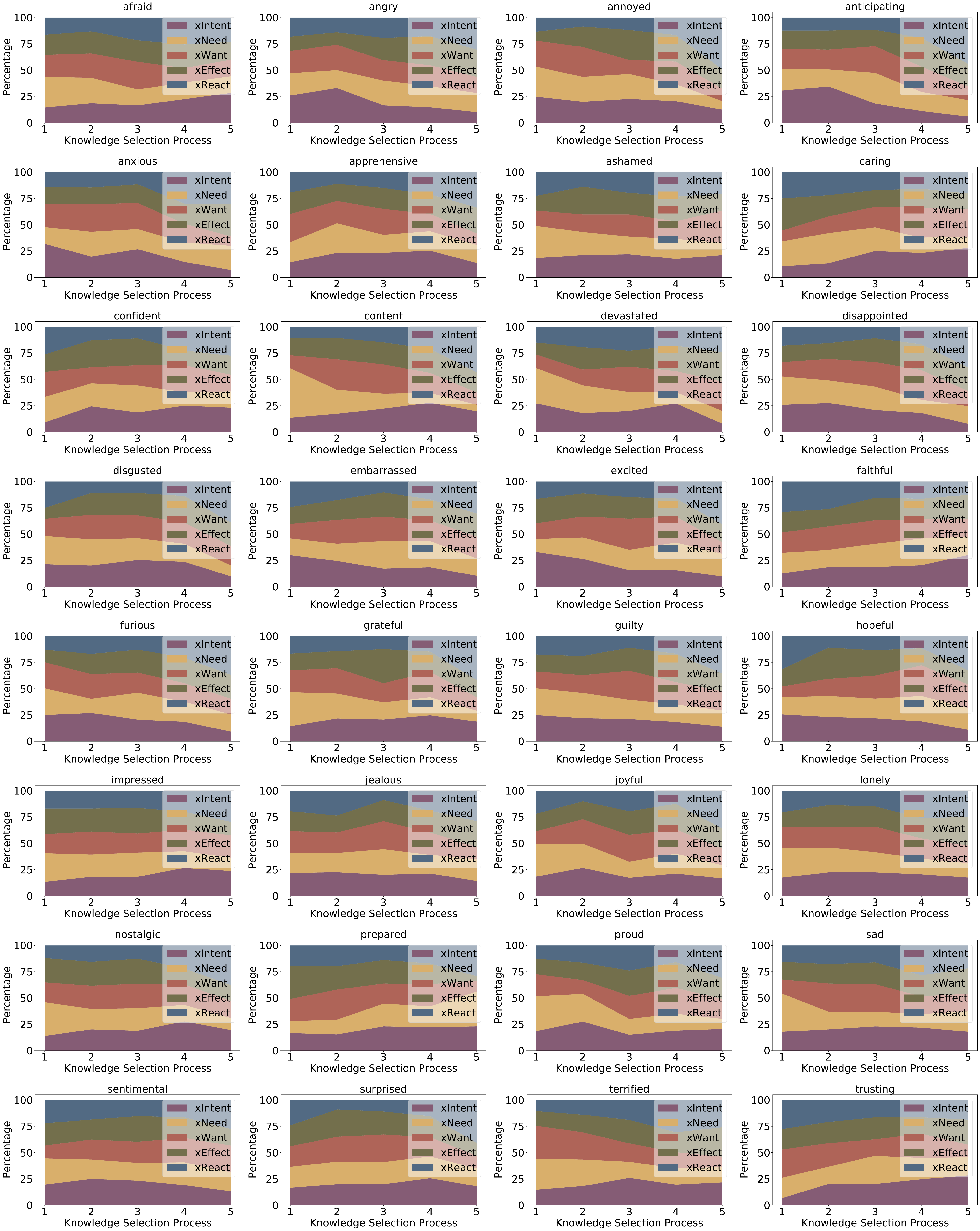}
\caption{Stackplot of the knowledge selection process.  }
\label{fig5}
\end{figure*}

We provide qualitative studies in Section 4.6. It includes 1) Ablation study of our cognition dynamics (Table \ref{tb:caseous});  2) Additional case study of generated responses from EmpatheticDiaglogues (Table \ref{tb:caseadd}); 3) Stackplot of the knowledge selection process for all the emotions in EmpatheticDiaglogues (Figure \ref{fig5}).

\begin{algorithm}[t]
\caption{Adaptive Knowledge Selection Method. }
\label{alg:algorithm}
\textbf{Input}: The set of knowledge representation $\mathbf{Z} = \{ \mathbf{Z}_r[0] \},  \mathbf{Z}_r[0] \in \mathbb{R}^d$, linear layer $g_{\theta}, \theta \in \mathbb{R}^{d \times q} $, the context representation vector $\mathbf{z}_{\textnormal{ctx}} \in \mathbb{R}^{d} $ from dialogue history encoder, the objective function of emotion classification $ \mathcal{L}_{\textnormal{emo}} $.

\begin{algorithmic}[0] 
\STATE \textbf{\textit{$\bullet$ Competition Stage}}:
\WHILE{ $len(\mathbf{Z}) > 1 $}
\STATE $m=1$
\STATE $I = \max_{ \mathbf{z} \in \mathbf{Z}}\{\mathcal{L_\textnormal{emo}}(g_{\theta}(\mathbf{z}),g_{\theta}(\mathbf{z}_{\textnormal{ctx}})) \}$
\STATE $\mathbf{f} = \{\mathcal{L_\textnormal{emo}}(g_{\theta}(\mathbf{z}),g_{\theta}(\mathbf{z}_{\textnormal{ctx}})), \mathbf{z} \in \mathbf{Z}  \} $
\STATE  ${\mathbf{G}}_m = \nabla_{{\bm{\theta}}} \mathbf{f}\in \mathbb R ^{d \times q} $
\STATE Solve Lagrange multiplier ${\bm{\lambda}} $:
\STATE  $  \min \limits_{\bm{\lambda}} \ (\frac12 {\bm{\lambda}} ^T {\mathbf{G}}_m{\mathbf{G}}_{m}^T {\bm{\lambda}} - \mathbf{f}^T {\bm{\lambda}})  \quad $
\STATE $s.t.\quad \sum^{J}_{i=1}\lambda_i =1,\lambda_i\geq 0.$
\IF{$ m = 1$}
\STATE ${\bm{\delta}}_m = -{\mathbf{G}}_{m}^T{\bm{\lambda}} $
\ELSE
\STATE $ {\bm{\delta}}_m = \bm{\delta}_{m-1} -{\mathbf{G}}_{m}^T{\bm{\lambda}} $
\ENDIF
\STATE $\mathbf{Z} = \mathbf{Z}_{- I} $  
\STATE $m = m + 1$
\ENDWHILE
\STATE \textbf{\textit{$\bullet$ Broadcasting Stage}}:
\STATE  $\widetilde{\mathbf{h}}_{k} = \mathbf{h}_{k} + \bm{\delta}_m   $
\end{algorithmic}
\end{algorithm}

\begin{table}[t]
\centering
\fontsize{10pt}{10pt}\selectfont
\begin{tabular}{@{}l|l@{}}
\toprule
\textbf{Context }        & \begin{tabular}[c]{@{}l@{}}Speaker: \\  
My friend borrowed quite  a \\lot   of money from me. I really \\ do   believe   he'll repay me. \end{tabular} \\ \midrule      
\textbf{Knowledge} & \begin{tabular}[c]{@{}l@{}}
xIntent: To be helpful \\
xWant: To repay the money \\
xNeed: To ask for a loan \\
xEffect: Gets a receipt \\
xReact: Happy; Relieved\end{tabular} 
\\ \midrule

\textbf{Emotion}            & \textbf{\textit{Trusting}} \\
\textbf{Selection Process} & \begin{tabular}[c]{@{}l@{}}xReact $ \rightarrow$ xEffect $ \rightarrow$ xIntent \\ $ \rightarrow$ xNeed $ \rightarrow$ xWant \end{tabular}\\

\textbf{Selected Knowledge} & xWant: \textbf{\textit{To repay the money}}

\\ \midrule  

\textbf{Ours (w/o A*)} & \begin{tabular}[c]{@{}l@{}}That is very nice of him. \\ What did  he do?   \end{tabular} \\
\textbf{Ours (w/ A*)} & I am sure he will repay you.    
\\ \midrule \midrule
\textbf{Context }        & \begin{tabular}[c]{@{}l@{}}Speaker: \\  
I had a nice meal and my \\ favorite beverage after work. \end{tabular} \\ \midrule      
\textbf{Knowledge} & \begin{tabular}[c]{@{}l@{}}
xIntent: To relax after work \\
xWant: To go to bed \\
xNeed: To go to the restaurant \\
xEffect: Has a full belly \\
xReact: Satisfied; Happy\end{tabular} 
\\ \midrule

\textbf{Content}            & \textbf{\textit{Trusting}} \\
\textbf{Selection Process} & \begin{tabular}[c]{@{}l@{}}xNeed $ \rightarrow$ xWant $ \rightarrow$ xReact \\ $ \rightarrow$ xEffect $ \rightarrow$ xIntent \end{tabular}\\

\textbf{Selected Knowledge} & xIntent: \textbf{\textit{To relax after work}}

\\ \midrule  

\textbf{Ours (w/o A*)} & \begin{tabular}[c]{@{}l@{}}What did you eat \end{tabular} \\
\textbf{Ours (w/ A*)} & \begin{tabular}[c]{@{}l@{}}Sounds like a relaxing day.\\ What did you drink?   \end{tabular} 
\\ \bottomrule
\end{tabular}
\caption{ Ablation study of our cognition dynamics.   A$\ast$ represents the adaptive knowledge selection method in the workspace module. Selection Process illustrates that at each the iteration, the most irrelevant knowledge is selected and eliminated.  }
\label{tb:caseous}
\end{table}


\begin{table*}[t]
\centering
\fontsize{10pt}{10pt}\selectfont
\begin{tabular}{@{}l|l@{}}
\toprule
\textbf{Context }        & \begin{tabular}[c]{@{}l@{}}Speaker: My family and I are going on vacation in a few weeks. We \\ rented a Largo Beachfront Condo  and I can not wait! \\
Listener: Oh I love the beach!! Which beach are you going to go to? \\
Speaker: We are going to be staying at Miramar Beach in Florida, about an hour \\ Panama City Beach.  The \textbf{\textit{view is amazing}} from our balcony and we are so ready!\end{tabular} \\ \midrule                                                                                                                                                                
\textbf{MoEL  }             & That is great! I hope you have fun!                                                                                                                                                                                                                                                                                                                                                                                   \\
\textbf{MIME   }            & That sounds like a lot of fun!                                                                                                                                                                                                                                                                                                                                             \\
\textbf{EmpDG  }          & That is awesome! I hope you are able to have a blast!                                                                                                                                                                                                                                                                                                                                                                             \\
\textbf{CEM}                & That sounds like fun!                                                                                                                                                                                                                                                                                                                                                                                    \\ \midrule
\textbf{Selection Process} & \begin{tabular}[c]{@{}l@{}}xIntend $ \rightarrow$ xEffect $ \rightarrow$ xReact $ \rightarrow$  xNeed $ \rightarrow$ xWant \end{tabular}\\

\textbf{Selected Knowledge} & xWant: \textbf{\textit{To enjoy the view}}; To enjoy the beach; To enjoy the day; To have fun                                                                                                                                                                                                                                                                                                                                      \\
\textbf{Emotion}            & \textbf{\textit{Anticipating}}                                                                                                                                                                                                                                                                                                                                                                                               \\
\textbf{Ours    }           & Sounds like a lot of fun. \colorbox{yellow}{I have never been there, but I will check it out!}                                                                                                                                                                                                                                                                                                                            \\ \midrule
\textbf{Gold}               & I bet it is tough. A regular income is nice but the freedom you get from working \\ for yourselfs is better.                                                                                                                                                                                                                                                                                                                                                                             \\ \midrule \midrule
\textbf{Context}            & 
\begin{tabular}[c]{@{}l@{}}Speaker: My parents keep taking my money  that is not nice of them. \\
Listener: Why do they keep taking it??  \\
Speaker: They need it for other stuff. \textbf{\textit{I never get a chance to save any of it}}.
\end{tabular}                                                                                                                                                                                                                                                                                                                \\ \midrule
\textbf{MoEL  }             & That is great! I hope you find a good job!                                                                                                                                                                                                                                                                                                                                             \\
\textbf{MIME  }             & That is a good thing to do.                                                                                                                                                                                                                                                                                                                                               \\
\textbf{EmpDG }             & That is a very nice thing to do.                                                                                                                                                                                                                                                                                                                                                                              \\
\textbf{CEM }               & Oh no! I am sorry to hear that.                                                                                                                                                                                                                                                                                                                                                                              \\ \midrule
\textbf{Selection Process} & \begin{tabular}[c]{@{}l@{}}xReact $ \rightarrow$ xEffect $ \rightarrow$ xIntent $ \rightarrow$  xWant $ \rightarrow$ xNeed \end{tabular}\\
\textbf{Selected Knowledge} & xNeed: \textbf{\textit{To get rid of it}}; To get rid of something; To buy it; To have it                                                                                                                                                                                                                                                                                                 \\
\textbf{Emotion}            & \textbf{\textit{Annoyed}}                                                                                                                                                                                                                                                                                                                                                                                             \\
\textbf{Ours}               & I am sorry to hear that. \colorbox{yellow}{Maybe you can talk to them about it.}                                                                                                                                                                                                                                                                                                                                                                      \\ \midrule
\textbf{Gold}               & \begin{tabular}[c]{@{}l@{}} Sorry to hear that. I found that it helps to have multiple savings accounts,\\  so whenever you get a certain amount of money ( like  5 ,  10 ) save it \\ and be strict not to spend it.\end{tabular}                                                                                                                                                                                                                                                                                                                                  \\ \midrule \midrule
\textbf{Context}            & \begin{tabular}[c]{@{}l@{}}Speaker: I went to visit a friend one day and her boyfriend made a pass at me. \\He actually   suddenly kissed me, as I was waiting for my friend to return. \\ I could not believe his nerve!  \\
Listener: Oh no, what did you do?  \\
Speaker: I quietly pushed him away, as I did not want my friend to hear. 
 \\ \textbf{\textit{I was so shocked and upset}}, 
since he knew I wouldn't want to make a scene \\and took advantage.\end{tabular} 

\\ \midrule
\textbf{MoEL  }             & That is so nice of her.                                                                                                                                                                                                                                                                                                                                              \\
\textbf{MIME   }            & That is nice of you.                                                                                                                                                                                                                                                                                                                                 \\
\textbf{EmpDG }             & That is a lot of people are not. I am sure you will be able to get another one.                                                                                                                                                                                                                                                                                                                                             \\
\textbf{CEM}                & That is terrible. Did you find out who did it?                                                                                                                                                                                                                                                                                                                                  \\ \midrule
\textbf{Selection Process} & \begin{tabular}[c]{@{}l@{}}xIntent $ \rightarrow$ xEffect $ \rightarrow$ xNeed $ \rightarrow$  xWant $ \rightarrow$ xReact \end{tabular}\\

\textbf{Selected Knowledge} & xReact: \textbf{\textit{Angry}}; Upset; Sad;  Annoyed;  Scared                                                                                                                                                                                                                                                                                                               \\
\textbf{Emotion}            & \textbf{\textit{Angry}  }                                                                                                                                                                                                                                                                                                                                                                                            \\
\textbf{Ours  }             & I would have \colorbox{yellow}{done the same thing.}                                                                                                                                                                                                                                                                                                                                            \\ \midrule
\textbf{Gold }              & Did you ever tell your friend?      \\   \midrule \midrule                                                                                                                                                 \textbf{Context}            & 
\begin{tabular}[c]{@{}l@{}}Speaker: My mother recently went to the bank for me. I love her so much. \\ 
Listener: That is really nice of her. It is nice to have someone \\ to help  you do things sometimes.  \\
Speaker: \textbf{\textit{Nice people are the best}}. Empathy is something we all need more of!
\end{tabular}                                                                                                                                                                                                                                                                                                                \\ \midrule
\textbf{MoEL  }             & That is very nice of you. It is always nice to have family.                                                                                                                                                                                                                                                                                                                                    \\
\textbf{MIME  }             & That is a good thing to do.                                                                                                                                                                                                                                                                                                                    \\
\textbf{EmpDG }             & That is a very nice thing to have a good mom.                                                                                                                                                                                                                                                                                                                                                            \\
\textbf{CEM }               & That is nice of you.                                                                                                                                                                                                                                                                                                                                                           \\ \midrule
\textbf{Selection Process} & \begin{tabular}[c]{@{}l@{}}xNeed $ \rightarrow$ xWant $ \rightarrow$ xIntent $ \rightarrow$  xReact $ \rightarrow$ xEffect \end{tabular}\\

\textbf{Selected Knowledge} & xEffect: \textbf{\textit{Is loved}}; Is respected; Is liked                                                                                                                                                                                                                                                                                                  \\
\textbf{Emotion}            & \textbf{\textit{Caring}}                                                                                                                                                                                                                                                                                                                                                                                             \\
\textbf{Ours}               & I agree with you there. It is nice \colorbox{yellow}{to have someone to lean on.}                                                                                                                                                                                                                                                                                                                                                                     \\ \midrule
\textbf{Gold}               & Yes. Too many people these days are mean to one another!                                                                                                                                                                                                                                                                                                                                                                                                                                                                                                                     

\\ \bottomrule
\end{tabular}
\caption{ Additional case study of generated responses from EmpatheticDiaglogues.    }
\label{tb:caseadd}
\end{table*}

\end{document}